\def\year{2019}\relax
\documentclass[letterpaper]{article} 
\usepackage{aaai}  
\usepackage{times}  
\usepackage{helvet}  
\usepackage{courier}  
\usepackage{url}  
\usepackage{graphicx}  
\usepackage{multirow}
\usepackage{tabularx}
\usepackage{amssymb}
\usepackage{amsmath}
\usepackage{caption}
\usepackage{verbatim}
\usepackage{flushend}

\nocopyright{}

\usepackage{adjustbox}
\usepackage{array}

\usepackage{color}
\newcommand{\todo}[1]{}
\renewcommand{\todo}[1]{{\color{red} TODO: {#1}}}

\newcolumntype{R}[2]{%
    >{\adjustbox{angle=#1,lap=\width-(#2)}\bgroup}%
    l%
    <{\egroup}%
}

\usepackage[utf8]{inputenc}
\DeclareUnicodeCharacter{193}{Á}

\title{Desiderata for Planning Systems in General-Purpose Service Robots}
\author{
Nick Walker\footnotemark[1]\textsuperscript{1}, Yuqian Jiang\thanks{Equal contribution.}\textsuperscript{2}, Maya Cakmak\textsuperscript{1}, Peter Stone\textsuperscript{2} \\
\textsuperscript{1}Paul G. Allen School of Computer Science \& Engineering, University of Washington, Seattle, USA\\
\textsuperscript{2}Department of Computer Science, University of Texas at Austin, Austin, USA\\
\{nswalker, mcakmak\}@cs.washington.edu\\
jiangyuqian@utexas.edu, pstone@cs.utexas.edu
}

\date{}

\begin{document}

\maketitle

\begin{abstract}
    General-purpose service robots are expected to undertake a broad range of tasks at the request of users. Knowledge representation and planning systems are essential to flexible autonomous robots, but the field lacks a unified perspective on which features are essential for general-purpose service robots. Progress towards planning and reasoning for general-purpose service robots is hindered by differing assumptions about users, the environment, and the overall robot system. In this position paper, we propose desiderata for planning and reasoning systems to promote general-purpose service robots. Each proposed item draws on our experience with research on service robots in the office and home and on the demands of these environments. Our desiderata emphasize support for natural human-interfaces as well as for robust fallback methods when interactions with humans and the environment fail. We highlight relevant work towards these goals.

\end{abstract}

\section{Introduction}

\begin{figure}
    \centering
    \includegraphics[width=\columnwidth]{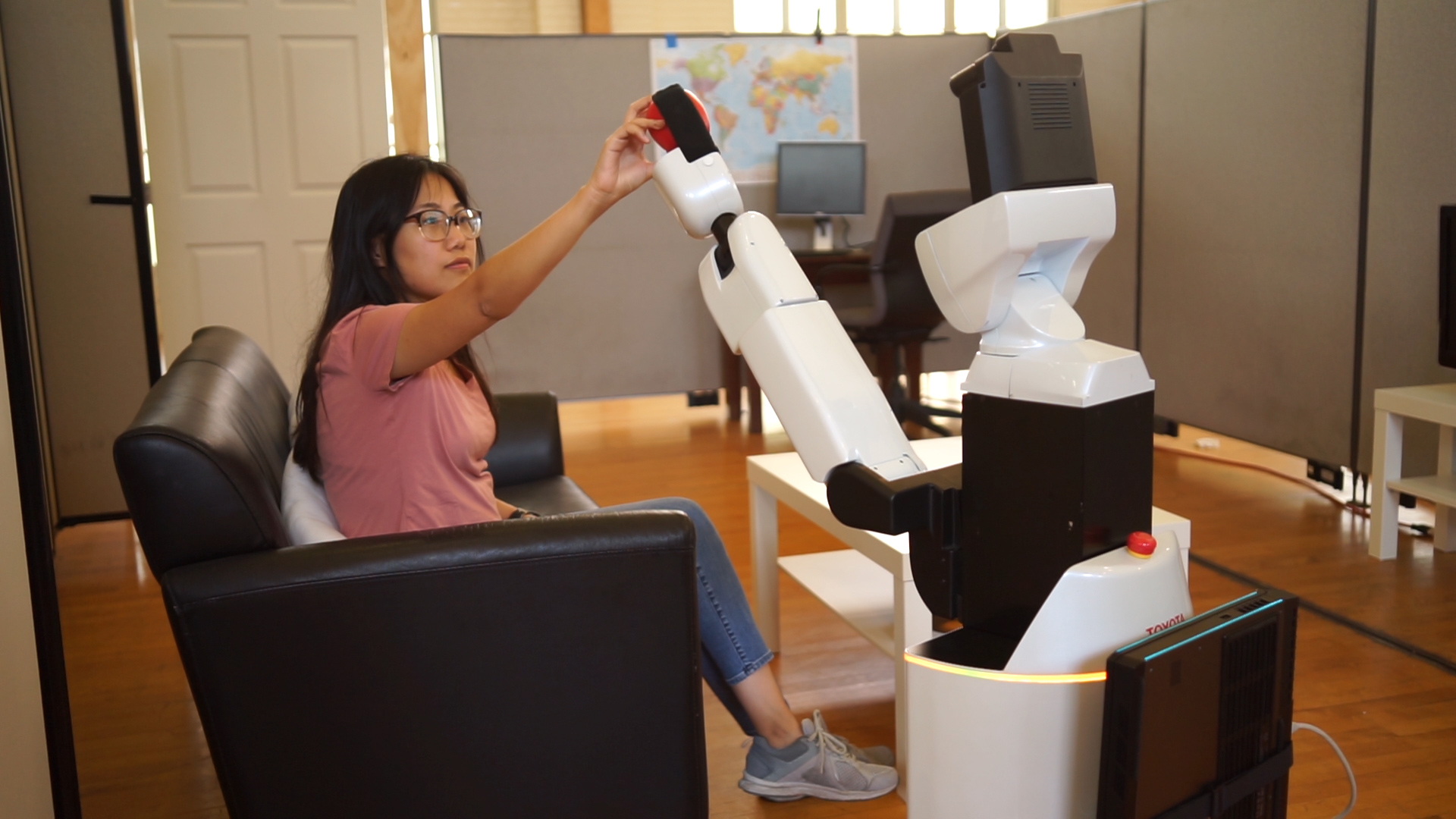}
    \caption{A Toyota Human Support Robot, representative of typical GPSRs, handing an apple to a user in a mock home environment.}
    \label{fig:handover}
\end{figure}

Creating a robot that is able to carry out a broad range of tasks in a household or office environment is a long-standing grand challenge for A.I. and robotics.  
Envisioned in science fiction for example as ``Rosie the Robot'' in the Jetsons, and the focus of competitions such as RoboCup@Home, we refer to such a futuristic embodied personal assistant as a \emph{General Purpose Service Robot} (GPSR). 
A GPSR, like the one shown in Figure~\ref{fig:handover}, integrates skills for navigating, observing and manipulating the environment, and interacting with users to service requests. 

The problem of coordinating low-level skills to accomplish complex goal-driven behavior is often addressed using AI planning techniques.
AI planning systems reason about a robot's feasible actions and generate a trajectory in an abstract action space based on models of the environment. 
These systems define structures that are used in all aspects of the robot's behavior including: the actions it can take and their effects on the world, the information it gathers, and attributes of the environment it perceives. 
For example, when a GPSR is asked to ``Grab a fruit from the kitchen," it might rely on its current knowledge about the layout of the environment and what objects are in the kitchen to produce a preliminary plan. 
As the GPSR executes the plan, its perception modules will store information in its representation of knowledge. 
If an action fails, the robot may invoke a reasoning process to infer the source of the failure, or to suggest alternative goals it could pursue. As this example illustrates, the knowledge representation and planning system play a critical role in orchestrating the complex behavior of a general-purpose robot. 
A well designed system could be beneficial to a range of tasks in various environments across different platforms.

Building a general-purpose service robot is a difficult task.
In practice, researchers focus their efforts on individual aspects of the general problem with the hope that progress on a part will contribute to progress on the whole. As a result, the community pursuing service robots has sought solutions to many different problems, at the expense of progress towards integrated general-purpose service robots.
Unlike other subsystems which are difficult to disentangle from hardware or particular environments, the knowledge representation and planning system of a service robot is a central and portable component that can be designed to address a wide set of requirements.
It can undergird progress towards more general robots.  

{}

 Three main factors give rise the constraints that GPSRs must operate in. Differing assumptions about these factors can motivate different designs of GPSR planning systems:

\begin{itemize}
    \item Users. Researchers make varying assumptions about how users will interact with robots. These motivate disparate approaches. What will users actually expect from systems? Can we derive from these a common set of functions that our systems should support?
    
    \item The environment. Expense and technical challenge makes it difficult to field robots in multiple environments. Often, researchers design systems to be flexible but rely on assumptions about the domains they are targeting. Which of these assumptions will still hold in other domains? Can we consolidate characteristics of various environments to better understand what features of planning systems support generalization across different domains?
    
    \item The robot system. Platforms are highly diverse in their design and capabilities, and it's rare that researchers have access to more than a handful. As a result, they may assume a level a of computation or a level of connectivity that doesn't hold with similar platforms. Can we extract the core set of hardware constraints that service robot planning systems should assume? 

\end{itemize}

 Progress towards the creation of planning systems for general-purpose service robots depends on our ability to agree on a basic set of goals. Therefore, an important task for the AI planning community should be understanding the requirements and constraints of general-purpose service robots. Towards this end, we propose desiderata for knowledge representation and reasoning systems in general-purpose service robots. Our purpose is to make explicit a position on how users, the environment, and the robot system drive the design of the knowledge representation and planning component. We hope that these points prompt further discussion and lay a path for work towards the creation of practical service robots in a wide range of environments.

\section{Desiderata}

A general-purpose service robot is a capable and intelligent helper. In the hospital, such a robot may be asked to fetch supplies. At home, a GPSR might set the table or store groceries. In the office, a GPSR might scout meeting rooms or locate coworkers. The key aspect of a GPSR is that its tasking is flexible. If a robot has some set of skills, we would designate it a general-purpose robot if it can sequence these skills to achieve more complex behavior, even when the particular goal has not been encountered before and the environment or context changes. Progress in individual capabilities like autonomous navigation and localization, object manipulation or visual recognition are all necessary to implement such a system, but in order to achieve generality, these components must operate at the behest of a larger planning system.

Researchers have long pursued general-purpose service robots. Since 2006, the international RoboCup@Home competition has challenged teams to field domestic service robots in a mock apartment. The competition tasks have included entertaining guests, cleaning, as well as an explicit \emph{General Purpose Service Robot} challenge, where the robot carries out tasks that an operator requests in spoken language \cite{Holz2013}. This task is designed to confront the robot with key challenges from the GPSR problem, such as interacting with the user to clarify a goal, planning without full knowledge of objects in the environment, and handling execution time failures. Success requires comprehensive integration of a robot planning and reasoning system with a real robot system that can operate under fixed time constraints. This task is perhaps the largest benchmarking effort for general-purpose service robots, but this vision is shared widely in the robotics research community \cite{alterovitz2016robot,Agostini2017,Torras2016}.

In the following, we describe some of the essential aspects of GPSRs and how they lead to our desiderata. 

\subsection{Users}

 Whether in a domestic, a medical, or a scholastic setting, GPSRs will interact primarily with non-expert users. These users may have detailed knowledge of the environment and their roles, but will have minimal familiarity with robots. This attribute drives several desiderata:

\begin{itemize}
    \item \emph{Human-interfaces}: GPSR planning systems must be able to accept commands and information from a wide variety of users. 
    Non-experts will not be able to provide input in planning systems' native representations, but will instead need to use special interfaces. 
    Such interfaces should facilitate all manner of user demands, including goals, constraints on planning and scheduling, and suggestions. Research towards the design of these interfaces continues, spanning natural language interaction~\cite{Cantrell2012,Thomason2019improving} to graphical user interfaces~\cite{Huang2017hri}. Planning systems must be designed to support one or more of these human-interfaces.
    
    \item \emph{Active error resolution}: Often, commands given by non-expert users cannot be formulated in the robot's representation because they are incomplete or ambiguous. Instead of discarding the command and requiring the user to start over, it's preferable to actively resolve these problems through dialog or other interactions with the user.
    
    \item \emph{Responsiveness}: Users will expect the robot to service commands promptly. Planning processes must not impede the robot's ability to quickly carry out tasks. If there are further interactions during the command taking phase, it is critical that queries can be answered within reasonable time to keep the users engaged. Even during execution, the system must always be ready to respond to interruptions from users asking questions or making new requests.
    
    \item \emph{Online adaptation}: User needs are diverse and change over time. As a consequence, a GPSR will be asked to carry out tasks for which it does not have adequate skills or planning knowledge. A GPSR's planning system must facilitate the acquisition of both new actions as well as the knowledge necessary to plan with these actions, either through end-user programming \cite{Huang2017hri}, interactive learning ~\cite{Laird2017}, or other methods.
    
    \item \emph{Explainability}: Explanations of the robot's plan can help build trust with the users. The planning system should be able to answer questions about the generated plans, and in case execution fails, explain what in the environment is different from the robot's model at planning time~\cite{Fox17explainable}. 

\end{itemize}

\subsection{The Environment}

We expect that GPSRs will find use in a wide range of environments, but a constant across these is the presence of people. Beyond handling robot-to-operator interactions, the robot must be able to cope with the demands of human-populated environments. This leads to additional desiderata:

\begin{itemize}

    \item \emph{Interpretability}: While people are skilled at perceiving each other's gross intentions, they are not generally capable of discerning the intentions of robots. Robots carrying out complex behaviors autonomously for long durations will create uncertainty in passersby. In addition to being friendly to direct users, a GPSR planning and reasoning system must include transparency mechanisms that support producing interpretable plans to build trust and minimize disturbance to humans in the environment.
    
    \item \emph{Ethical}: 
    GPSRs must respect the values of the people around them.
    A recent European Commission report identified respect for human autonomy, prevention of harm, fairness and explicability as the core ethical principles that trustworthy AI systems must uphold~(\citeyear{EU2019}).
    Balancing adherence to ethical concerns in general will be a challenging task, so GPSR planning and reasoning systems must support straightforward encoding and tuning of these principles.
    
    \item \emph{Safety}: GPSRs will be expected to guarantee that their actions bring no harm to the humans in their environment.
    Ensuring safety is challenging for any autonomous robot, but is particularly difficult given the dynamic nature of a GPSR's tasking.
    How can planning systems help ensure that the robot only undertakes safe tasks?
    
    \item \emph{Resource efficiency}: GPSRs will operate in domains that have inherent resource constraints, such as the number of elevators, or the available supply of human patience. Planning systems that interact with these limited resources must model and optimize for the costs of the robot's actions.
    
    \item \emph{Handle uncertainty}: Since GPSRs operate in highly dynamic environments, the robot's model at planning time often turns out to be inaccurate during execution. 
    The planning system should therefore be able to robustly handle these uncertainties at planning time or through replanning at execution time.
    
    \item \emph{Adaptability}: Some deviations in a GPSR's environment will represent a major or lasting change. 
    Beyond simply tolerating these at execution time, the planning system should incorporate this information and not generate plans that are likely to fail given the current state of the environment. 
    Rather, the planning system should generate better initial plans that adapt to new situations, based on recent experience.
    
    \item \emph{Open-world}: The real world for a GPSR is ``open''; the robot's knowledge base cannot capture the full state of the environment. 
    Certain information may not be available to a GPSR at planning time, such as the current location of an object that the user requested. 
    Therefore, the planning system should not make the ``closed-world'' assumption and should be capable of planning for knowledge acquisition.
\end{itemize}

\subsection{Robot System}

A robot is a complex combination of hardware and software components, and each is unique in its set of tradeoffs. 
Planning systems operate atop representations that are notionally abstract to the particulars of a platform, however, in practice, the overall robot system still imposes important constraints on the design of the planning system. 
Work towards practical planning systems for GPSRs must strive for:

\begin{itemize}
    \item \emph{Robustness}: An unexpected failure in a GPSR's planning system will most immediately erode user trust and faith. 
    Further, inopportune issues have the potential to endanger the robot or its surroundings, and the long duration of deployments makes such occurrences almost unavoidably probable if not adequately addressed.
    Planning and reasoning systems for GPSRs must achieve high standards of robustness, making them a prime target for formal verification efforts. 
    
    \item \emph{Portability}: Robots have unique strengths and weaknesses, but the core set of necessary capabilities for GPSRs are fixed; the robot will have to navigate through and manipulate its environment, as well as interact with people. To enable portability, planning systems should model at least this minimum set of capabilities, regardless of the details of particular platforms. 
    
    \item \emph{Compute efficiency}: Mobile robots are constrained by battery technology. In practice, the majority of a GPSR's power budget is spent on computation, which leads robot designers to adopt limited, power efficient compute solutions. As a result, GPSR planning systems must be able operate within a limited compute budget.
    
    \item \emph{Connectivity tolerance}: Network reliability in robotics laboratory environments is an upper bound on the conditions that real GPSRs will encounter when deployed. Though standards and hardware for consumer grade networking continue to improve, robots' preferred bandwidth and signal characteristics will not be well served by typical equipment for some time. As a result, practical GPSR planning systems cannot assume reliable network connectivity.

\end{itemize}

\section{Recent Directions}

There is still a significant gap between existing systems and the fulfillment of the desiderata. 
Taking as an example work from Khandelwal et. al. (\citeyear{Khandelwal2017bwibots}) and Thomason et. al. (\citeyear{Thomason2019improving}), it is possible to achieve high levels of responsiveness and support dialog interactions in a deterministic classical planning regime.
The efficiency enabled by this paradigm however comes at the expense of rich expression of the probabilistic aspects of the environment. 
The authors could have pursued probabilistic systems that have straightforward means for expressing the openness of the world or the uncertainty of the robot's knowledge, but with current methods, they would have necessarily found the responsiveness of the resulting system limited. 

Besides difficulties in integrating existing technologies, fulfillment of the desiderata is challenging because some items have received less attention from the planning community and have few mature approaches.
In the remainder of this section, we highlight directions where there is ongoing work and opportunities for progress towards our desiderata.
For a comprehensive survey on advances in robot planning technologies we refer the reader to the recent survey by Ingrand and Ghallab~(\citeyear{Ingrand2017}).

There is growing interest in human-aware and explainable planning \cite{sreedharan2017balancing}. 
Recent progress includes new problem formulations, metrics, and initial approaches in generating explicable plans \cite{Zhang2017} and plan explanations \cite{Chakraborti2017}. 
Recent works have advanced techniques for learning and refining planning operators \cite{Agostini2017}, a critical component in interactive learning. 
Efforts to refine planning components in cognitive architectures are creating some of the first integrated interactive learners \cite{Mininger2016interactively}.

Planning under uncertainty is attacked with many different methods. 
There have been recent efforts to combine logical and probabilistic planning and reasoning methods in integrated architectures for robots \cite{Sridharan2018reba}.
A state-of-the-art integrated robot task planning system by Hanheide et al. (\citeyear{Hanheide2017}) leverages a probabilistic planner to tackle the issues of reasoning in an open and uncertain world, as well as interactive goal acquisition and failure explanation. 
Recent work approaches adaptive planning by feeding information gathered during plan execution, such as action costs and user preferences, back to the planning system \cite{jiang2018integrating,Wilde2018learning}.
Other methods for combining learning and planning have demonstrated that deliberative controllers can enhance the exploration capabilities of deep reinforcement learning agents in complex visual simulators~\cite{Gordon2019}.

The importance of speed and responsiveness are recognized in the planning community, and benchmarked in the International Planning Competition \cite{lopez2015ipc,Vallati2015}. 
Some researchers are beginning to consider how to design systems that work across related domains and platforms \cite{Lima2018IntegratingCP,hart18fss}. 
Numerous other deployed systems with planning capabilities highlight enthusiasm for creating robust systems that can operate autonomously for long durations in real environments \cite{Veloso2015,Hawes2017STRANDS,Khandelwal2017bwibots}.

\section{Conclusion}

The objectives laid out in this paper are challenging, but they also present a unique opportunity for collaboration in the planning and robotics communities. Research that advances any of the desiderata is valuable, but progress towards general-purpose service robots is enhanced when we actively consider how our contributions interface with an integrated system. We hope that by laying out the most important objectives for a general-purpose service robot, the community will not only pursue each direction as an intriguing problem on its own, but also develop new approaches and integrated systems that aim to address the full set of desiderata.

\bibliographystyle{AAAI}
\bibliography{references_manual}
\end{document}